\documentclass[10pt,twocolumn,letterpaper]{article}

\usepackage{iccv}
\usepackage{times}
\usepackage{epsfig}
\usepackage{graphicx}
\usepackage{amsmath}
\usepackage{amssymb}
\usepackage{stmaryrd}
\usepackage{yhmath}

\usepackage{booktabs} 
\usepackage{multirow}
\usepackage{subcaption}

\usepackage{soul}
\usepackage[dvipsnames]{xcolor}

\usepackage[pagebackref=true,breaklinks=true,letterpaper=true,colorlinks,bookmarks=false]{hyperref}

\iccvfinalcopy 

\def\iccvPaperID{6964} 

\ificcvfinal\fi

\usepackage{amsmath,amsfonts,bm}

\def\eqref#1{equation~\ref{#1}}

\def\1{\bm{1}}

\def\rmU{{\mathbf{U}}}

\def\rmW{{\mathbf{W}}}

\def\ve{{\bm{e}}}

\def\vk{{\bm{k}}}

\def\vn{{\bm{n}}}

\def\vq{{\bm{q}}}

\def\vv{{\bm{v}}}

\def\vx{{\bm{x}}}

\def\vz{{\bm{z}}}

\def\mE{{\bm{E}}}

\def\mN{{\bm{N}}}

\def\mR{{\bm{R}}}

\def\mT{{\bm{T}}}

\def\mV{{\bm{V}}}

\def\mZ{{\bm{Z}}}

\DeclareMathAlphabet{\mathsfit}{\encodingdefault}{\sfdefault}{m}{sl}
\SetMathAlphabet{\mathsfit}{bold}{\encodingdefault}{\sfdefault}{bx}{n}

\def\gE{{\mathcal{E}}}

\def\gV{{\mathcal{V}}}

\def\gX{{\mathcal{X}}}

\def\sR{{\mathbb{R}}}

\newcommand{\R}{\mathbb{R}}

\DeclareMathOperator*{\argmax}{arg\,max}

\usepackage{enumitem}
\usepackage[numbers]{natbib}
\usepackage{colortbl}

\usepackage{multicol}

\renewcommand{\paragraph}[1]{\vspace{0em}\noindent\textbf{#1}.}

\usepackage{color}
\usepackage{xcolor}
\definecolor{turquoise}{cmyk}{0.65,0,0.1,0.3}
\definecolor{purple}{rgb}{0.65,0,0.65}
\definecolor{dark_green}{rgb}{0, 0.5, 0}
\definecolor{orange}{rgb}{0.8, 0.6, 0.2}
\definecolor{red}{rgb}{0.8, 0.2, 0.2}
\definecolor{darkred}{rgb}{0.6, 0.1, 0.05}
\definecolor{blueish}{rgb}{0.0, 0.3, .6}
\definecolor{light_gray}{rgb}{0.7, 0.7, .7}
\definecolor{pink}{rgb}{1, 0, 1}
\definecolor{greyblue}{rgb}{0.25, 0.25, 1}
\definecolor{tab_blue}{HTML}{1f77b4}
\definecolor{tab_orange}{HTML}{ff7f0e}
\definecolor{LightRed}{rgb}{0.99,0.89,0.89}

\definecolor{mesh_misty_rose}{HTML}{e6aaa3}
\definecolor{mesh_yellow}{HTML}{ffba00}

\newcommand{\Fig}[1]{Fig.~\ref{fig:#1}}
\newcommand{\Figure}[1]{Figure~\ref{fig:#1}}

\newcommand{\eq}[1]{(\ref{eq:#1})}

\newcommand{\Section}[1]{Section~\ref{sec:#1}}

\renewcommand{\paragraph}[1]{\vspace{.2em}\noindent\textbf{#1}.}

\newcommand{\SupplementaryMaterial}{{\color{darkred} supplementary material}\xspace}

\begin{document}
\newcommand{\yueqi}[1]{{\color{magenta}[Yueqi: #1]}}
\newcommand{\leo}[1]{{\color{red}[Leo: #1]}}
\newcommand\orl[1]{\textcolor{blue}{(\textbf{Or}: #1)}}

\title{Vector Neurons: A General Framework for SO(3)-Equivariant Networks}

\author{Congyue Deng$^1$ \quad Or Litany$^2$ \quad Yueqi Duan$^1$ \quad Adrien Poulenard$^1$ \\
Andrea Tagliasacchi$^{3,4}$ \quad Leonidas Guibas$^1$ \\
$^1$Stanford University \quad $^2$NVIDIA  \quad $^3$Google Research \quad $^4$University of Toronto \\
}

\maketitle
\ificcvfinal\thispagestyle{empty}\fi

\begin{abstract}
Invariance and equivariance to the rotation group have been widely discussed in the 3D deep learning community for pointclouds. Yet most proposed methods either use complex mathematical tools that may limit their accessibility, or are tied to specific input data types and network architectures.
In this paper, we introduce a general framework built on top of what we call \emph{Vector Neuron} representations for creating SO(3)-equivariant neural networks for pointcloud processing.
Extending neurons from 1D scalars to 3D vectors, our vector neurons enable a simple mapping of SO(3) actions to latent spaces thereby providing a framework for building equivariance in common neural operations -- including linear layers, non-linearities, pooling, and normalizations.
Due to their simplicity, vector neurons are versatile and, as we demonstrate, can be incorporated into diverse network architecture backbones, allowing them to process geometry inputs in arbitrary poses.
Despite its simplicity, our method performs comparably well in accuracy and generalization with other more complex and specialized state-of-the-art methods on classification and segmentation tasks. We also show for the first time a rotation equivariant reconstruction network. 
Source code is available at \url{https://github.com/FlyingGiraffe/vnn}.

\end{abstract}

\section{Introduction}

With the proliferation of lower-cost depth sensors, learning on 3D data has seen rapid progress in recent years. Of particular interest are pointcloud networks, such as PointNet~\cite{qi2017pointnet} or ACNe~\cite{sun2020acne} that fully respect the inherent set symmetry -- that point sets are not ordered -- by incorporating order-invariant and/or order-equivariant layers.
Yet, there are other important symmetries that have been less perfectly addressed in the context of pointcloud processing, with 3D rotations being a prime example. Consider a scenario where one scans an object using their LIDAR-equipped phone to retrieve similar objects. Clearly, the global object pose should not affect the query result. PointNet uses spatial transformer layers \cite{jaderberg2015spatial}, which only attain approximate pose invariance while also requiring extensive augmentation at train time. 

To avoid an exhaustive data augmentation with all possible rotations, there is a need for network layers that are equivariant to both order and SO(3) symmetries. Recently, two approaches have been introduced to tackle this setting: Tensor Field Networks \cite{thomas2018tensor} and SE(3)-Transformers \cite{fuchs2020se}. While guaranteeing equivariance by construction, both frameworks involve an intricate formulation and are hard to incorporate into existing pipelines as they are restricted to convolutions and rely on relative positions of adjacent points.

\begin{figure}
\centering
\includegraphics[width=.85\linewidth]{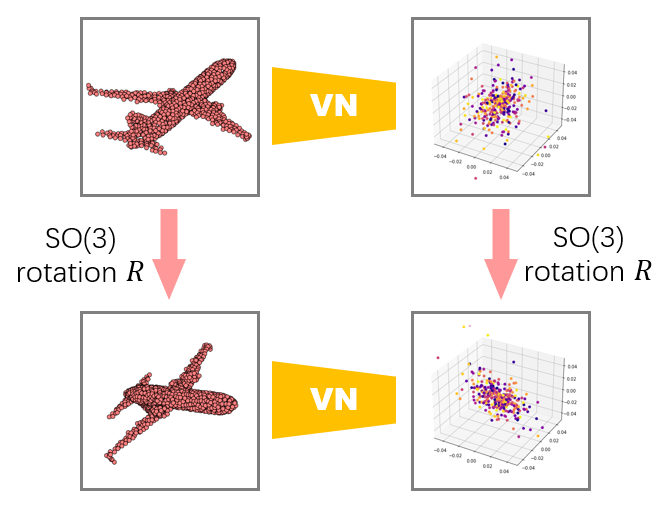}
\caption{By lifting latent representations from vectors of scalar entries to vectors of 3D points (i.e., matrices) we facilitate the creation of a simple rotation equivariant toolbox allowing the implementation of fully equivariant pointcloud networks.}
\label{fig:teaser}
\end{figure}
\begin{figure}
\centering
\includegraphics[width=.85\linewidth]{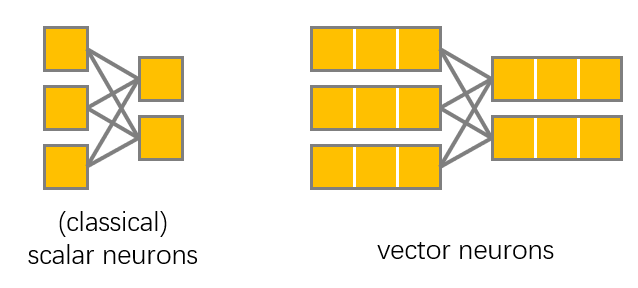}
\caption{\textbf{Linear layer --}
Typical neural networks today are built with ``scalar'' neurons -- where the output of the non-linearities in a given layer is an ordered list of \textit{scalars}. We extend deep networks to allow for ``vector'' neurons -- where the output of the non-linearity is an ordered list of \textit{vectors}.}
\label{fig:linear_layer}
\end{figure}

In this work, we address these issues by proposing a simple, lightweight framework to build SO(3) equivariant and invariant pointcloud networks. A core ingredient in our framework is a \textit{Vector Neuron} (VN) representation, extending classical scalar 
neurons to 3D vectors. Consequently, instead of   latent vector representations which can be views as ordered sequences of scalars, we deploy latent matrix representations which can be viewed as (ordered) sequences of 3-vectors. Such a representation supports a direct mapping of rotations applied to the input pointcloud to intermediate layers. This is in contrast to more complex solutions based on Wigner D-matrices \cite{cohen2018spherical}. Another appealing property of VN representations is that they remain equivariant to linear layers by construction. The challenge in building a fully-equivariant network lies in the  non-linear activations. In particular, standard neuron-wise activation functions such as ReLU will not commute with a rotation operation. A key contribution in this work is a 
3D generalization of classical activation functions by implementing them through a learned direction. For example, when applied to a vector neuron, a standard fixed direction ReLU activation would simply truncate the half-plane that points in its opposite direction. Instead, dynamically predicting an activation direction in a linear data-dependent fashion allows us to guarantee equivariance. We further provide an invariant pooling operation as well as normalization layers, which altogether render our framework compatible with various pointcloud network backbones. To demonstrate its versatility and efficiency, we implemented vector neuron versions of two popular architectures: PointNet and DGCNN, and tested them on three different downstream tasks: 
classification~(permutation invariant and rotation invariant), segmentation~(permutation equivariant and rotation invariant), and reconstruction~(rotation equivariant on the encoder side, and rotation invariant on the decoder side).
Despite its simplicity and lightweight architecture, in all tasks, our VN achieved top performance when tested on randomly rotated shapes compared with other equivariant architectures, and markedly improved performance compared to augmentation-induced equivariance approaches.

\vspace{.5em}
\noindent
To summarize, our key contributions are:
\vspace{-.5em}
\begin{itemize}[leftmargin=*]
\setlength\itemsep{-.3em}
\item We propose a new versatile framework for constructing SO(3)-equivariant pointcloud networks. 
\item Our building blocks are lightweight in terms of the number of learnable parameters and can be easily incorporated into existing network architectures.
\item We support a variety of learning tasks, in particular, we are the first to demonstrate a 3D equivariant network for 3D reconstruction.
\item When evaluated on classification and segmentation, our VN version of popular non-equivariant architectures achieve state-of-the-art performance.  
\end{itemize}
%

\section{Related Work}
The lack of robustness to rotation of classical deep learning architectures for pointcloud processing like PointNet~\cite{qi2017pointnet}, PointNet++~\cite{qi2017pointnetpp}, Dynamic Graph CNN (DGCNN)~\cite{wang2019dynamic}, PCNN~\cite{atzmon2018point}, PointCNN~\cite{li2018pointcnn}~(and many others) has driven interest for rotation invariant and equivariant designs.
In recent years the field of rotation invariant and equivariant deep learning for geometry processing has been rapidly developing. 
In what follows, we briefly review methods that achieve invariance and equivariance, as well as overview those that achieve equivariance via pose estimation.

\paragraph{Rotation invariant methods}
Rotation invariance is a desirable property for tasks like shape classification or segmentation. Many rotation invariant architectures 
\cite{liu2018deep, poulenard2019effective, chen2019clusternet, zhang2019rotation, zhang2020global, li2020rotation, zhao2019rotation, rao2019spherical} have been proposed to address these issues. For example, \cite{chen2019clusternet, zhang2019rotation, zhang2020global, li2020rotation} introduce cleverly designed rotation invariant operations. GC-Conv \cite{zhang2020global} relies on multi-scale reference frames based on PCA. RI-Framework \cite{li2020rotation} and LGR-Net~\cite{zhao2019rotation} pairs local invariant information with global context. Some works like LGR-Net \cite{zhao2019rotation} use surface normals in addition to the points coordinates. SFCNN \cite{rao2019spherical} proposes an approach similar to multi-view by mapping input pointclouds to a sphere and performing operations on the sphere. Other works like \cite{liu2018deep, poulenard2019effective} rely on more principled approaches borrowing tools from equivariant deep learning.

\paragraph{Rotation equivariant methods}
Recently multiple rotation equivariant deep learning architectures have emerged. A whole body of work is built on the theory of $\mathrm{SO}(3)$ representations \cite{thomas2018tensor, kondor2018clebsch, esteves2018learning, weiler20183d, anderson2019cormorant} -- most of these works rely on the concept of convolution with steerable kernel bases. A steerable kernel basis is a family of function undergoing a rotation in function space given a rotation of their input parameter. Features computed through these convolution inherit this equivariant behavior. A rotation of the object in euclidean space induces a rotation of the features in feature space. We refer to \cite{lang2020wigner} for a general theory of steerable kernels. Other works like EMVnet \cite{esteves2019equivariant} consider a multi-view image based representation of the shapes based on renderings of meshes. In the context of pointcloud network, the universality of rotation equivariance has been studied in \cite{dym2020universality}.

\paragraph{Equivariance via pose estimation}
\citet{qi2017pointnet} achieved approximate pose equivariance by factoring out SO(3) transformations through object pose estimation.
Most works in the literature study \textit{instance-level} pose estimation, where the ground-truth canonical pose of the 3D CAD models corresponding to the input pointcloud is available~\cite{brachmann2014learning}.
More recently \citet{wang2019normalized} introduced \textit{category-level} pose estimation, and extension to articulated objects has also been proposed~\cite{li2020categorylevel}.
While both these methods~\cite{wang2019normalized,li2020categorylevel} need explicit 2D-to-3D supervision, relaxing supervision is possible by borrowing ideas from Transforming~Auto-Encoders~\cite{hinton2011transforming,unsupervised}.
However, while \citet{sun2020caca} learn \textit{category-level} as well as \textit{multi-category} pose estimation in a fully unsupervised fashion, the underlying equivariant backbone~\cite{sun2020acne} is only equivariant by \textit{augmentation}.

\section{Method}
\label{sec:method}
We introduce \emph{Vector Neuron  Networks} (VNNs), a straightforward extension to classical ReLU networks that provides $\mathrm{SO}(3)$ equivariance by construction.
Neurons in standard artificial neural networks are built from scalars $z \in \R$. When stacked into an ordered list, these neurons form a $C^{(d)}$ dimensional latent feature $\vz {=} [z_1, z_2, \cdots, z_{C^{(d)}}]^\top\in \R^{C^{(d)}}$, where $(d)$ indexes the layer depth\footnote{For ease of notation, in what follows we will remove the layer index~$d$ whenever it is clear from context keeping in mind that the operations we introduce are per-layer.}.

However, when processing data embedded in $\R^3$ like 3D pointclouds, realizing the effect of SO(3) transformations applied to the input shape on these vector hidden layers is not obvious.
In particular, here we are interested in constructing  rotation-equivariant learnable layers, namely layers that commute with the action of the rotation group.

To this end, we propose to~``lift'' the neuron representation from a scalar $z \in \R$ to a vector $\vv \in \R^3$, leading to what we call a Vector Neuron (VN).
This results in list of Vector Neurons (matrix) 
$\mV {=} [\vv_1,\vv_2,\cdots,\vv_C]^\top \in \R^{C\times3}$.
Similar to standard latent representations, this vector-list feature can be used to encode an entire 3D shape, part of it, or a single point. In particular, when representing an (order-less) set of $N$ points $\gX = \{\vx_1,\vx_2,\cdots,\vx_N\} \in \sR^{N \times 3}$ in a pointcloud we can consider a collection of $N$ such vector-list features $\gV = \{\mV_1,\mV_2,\cdots,\mV_N\} \in \sR^{N \times C \times 3}$. 
Similar to standard neural networks, the number of latent channels $C^{(d)}$ can change between layers via a mapping:
\begin{equation}
    \gV^{(d+1)} = f(\gV^{(d)};\theta): \R^{N\times C^{(d)} \times 3} \to \R^{N\times C^{(d+1)} \times 3} ,
\label{eq:vn_mapping}
\end{equation}
where $\theta$ represents learnable parameters. 

Using this representation, the action of a rotation is made simple to implement, as it involves the application of a standard rotation matrix. Critically, we require that the mapping satisfies rotation equivariance, namely, for any rotation matrix $\mR \in \mathrm{SO}(3)$:
\begin{equation}
    f(\gV R;\theta) = f(\gV;\theta) R,
\end{equation}
where we interpret the application of the rotation matrix to the set as $\gV R = \{\mV_i R\}_{i=1}^N$. 
%
To facilitate equivariance in standard pointcloud network architectures, we construct VN layers following traditional designs via a combination of a linear map (\Section{linear}) followed by a per-neuron non-linearity (\Section{nonlinearity}). We additionally introduce equivariant pooling (\Section{pooling}) and normalization layers (\Section{normalization}).
With these building blocks we are able to assemble a rich variety of complex neural networks in equivariance, including the most basic VN Multi-Layer Perceptron (VN-MLP) as a sequence of alternating linear and non-linear layers.

\subsection{Linear layers -- \Figure{linear_layer}}
\label{sec:linear}

We begin by realizing the mapping $f$ introduced in equation \ref{eq:vn_mapping} as a linear operator -- a fundamental module of neural networks. 
Given a weight matrix $\rmW \in \R^{C'\times C}$, we define a linear operation $f_{\text{lin}}(\cdot; \rmW)$ acting on a vector-list feature $\mV \in \gV \in\sR^{N \times C \times 3}$ as follows: 
\begin{equation}
    \mV' = f_{\text{lin}}(\mV;\rmW) = \rmW \mV \in \R^{C'\times3}.
\end{equation}
%
We verify that a rotation matrix $R \in \mathrm{SO}(3)$ commutes with this linear layer:
\begin{equation}
    f_{\text{lin}}(\mV R;\rmW) = \rmW\mV R = f_{\text{lin}}(\mV;\rmW) R = \mV' R,
\label{eq:linequivariance}
\end{equation}
yielding the desired equivariance property.
Note that we omit a bias term as an addition of a constant vector that would interfere with equivariance. 
Further, note that while this layer is SO(3) equivariant, we can achieve SE(3) equivariance by centering $\mV$ at the origin.
Finally, depending on the setting, $\rmW$ may or may not be shared across the elements $\mV$ of $\gV$. 
%

\begin{figure}[t]
\centering
\includegraphics[width=.78\linewidth]{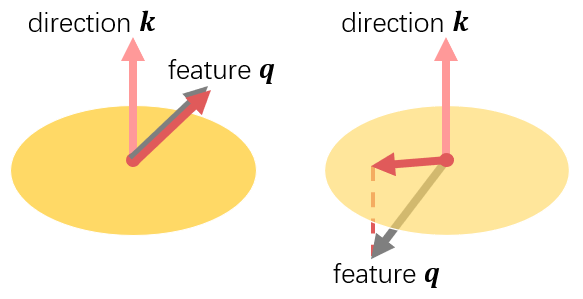}
\caption{\textbf{Non-linearity --}
Our non-linearity generalizes ReLU by acting on vectors rather than scalar inputs, and is parametric with respect to a learned direction $\vk$:
(left) when the input feature $\vq$ lies in the half-space defined by $\vk$, the feature stays unchanged;
(right) when the input feature $\vq$ lies in the half-space defined by $-\vk$, the feature component in that half-space is clipped. See equation \ref{eq:VN-ReLU}.
} 
\label{fig:nonlinearity}
\end{figure}

\begin{figure}[t]
\centering
\includegraphics[width=.95\linewidth]{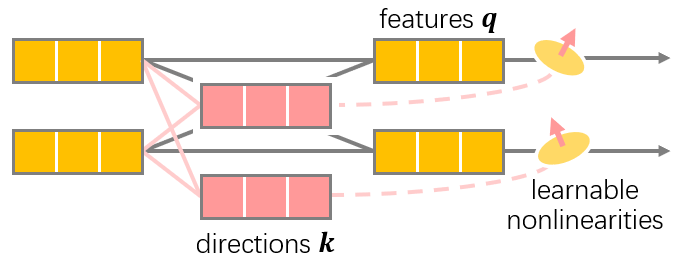}
\caption{\textbf{Non-linear layer --}
Our VN non-linearity is parametric in a learned direction $\vk$, derived from the input features via a learnable linear layer.
%
}
\label{fig:nonlinearity_overall}
\end{figure}
\subsection{Non-linear layers -- \Figure{nonlinearity} and \Figure{nonlinearity_overall}}
\label{sec:nonlinearity}
Per-neuron non-linearity is key to the representation power of neural networks.
As evident from recent literature, especially useful are functions that split the input domain into two half spaces and map them differently~(e.g.~ReLU, leaky-ReLU, ELU, etc.).
In the case of VN, a 3D version of these non-linearities, $\mV' = f_\text{ReLU}(\mV)$, is needed. 
Yet, committing to a fixed frame (i.e., one that does not depend on the input pose) like the standard coordinate system would violate equivariance.
Instead, we propose to dynamically predict a direction from the input vector-list feature. We then generalize the classical ReLU by truncating the portion of a vector that points into the negative half-space of the learned direction.
%

More formally, given an input vector-list feature $\mV \in \R^{C\times3}$, for each output vector-neuron $\vv' \in \mV'$ we learn \emph{two} weight matrices $\rmW \in \R^{1\times C}$ and $\rmU \in \R^{1\times C}$, linearly mapping the input feature $\mV$ to a feature $\vq \in \R^{1\times 3}$ and a direction $\vk \in \R^{1\times 3}$:
\begin{equation}
    \vq = \rmW\mV, \quad \vk = \rmU\mV\,.
    \label{eq:linearEquivariance}
\end{equation}
We then define the output VN as:
\begin{equation}
    \vv'
    = \begin{cases}
        \vq &
        \text{if\ } \langle \vq,\vk \rangle \geqslant 0 \\
        \vq - \left\langle \vq,\frac{\vk}{\|\vk\|} \right\rangle \frac{\vk}{\|\vk\|} &
        \text{otherwise,}
    \end{cases}
    \label{eq:VN-ReLU}
\end{equation}
resulting in an output vector-list: $f_\text{ReLU}(\mV) = [\vv']_{c=1}^{C}$
\footnote{In practice, when computing for the unit direction vector $\vk/\|\vk\|$ we implement $\vk/(\|\vk\|+\varepsilon)$ with a small margin $\varepsilon$ in the denominator to avoid division by zero at the origin}.

As illustrated in \Figure{nonlinearity}, $\vq$ can be decomposed into two components: $\vq_\parallel$ and $\vq_\perp$ that are parallel and orthogonal to $\vk$, respectively. Analogous to the standard scalar ReLU, we apply the nonlinear function to $\vq_\parallel$ along the direction $\vk$ by clipping $\vq_\parallel$ to zero, while keeping $\vq_\perp$ unchanged.
Other types of split-case functions (e.g. leaky-ReLU) follow immediately from this definition. We discuss these and other types of non-linearities in the \SupplementaryMaterial.

It is easy to verify that $f_\text{ReLU}$ is rotation equivariant. In particular, both $\vq$ and $\vk$ are linear maps of $\mV$ and thus commute with a rotation matrix as discussed in~\eq{linequivariance}.
Moreover, the inner-product term in the second case would cancel out an orthogonal matrix $\left\langle \vq \mR, \vk \mR \right\rangle = \left\langle \vq, \vk \right\rangle$ resulting in a scalar multiplication of a $\vk$, which is again equivariant.




\subsection{Pooling layers}
\label{sec:pooling}

Pooling is widely used when aggregating local/global neighbourhood information, either spatially (e.g. PointNet++) or by feature similarity (e.g. DGCNN). 
While mean pooling is a linear operation that respects rotation equivariance, we also define a VN max pooling layer as a counterpart to the classical max pooling on scalars.

For global pooling, we are given a set of vector-lists $\gV \in \sR^{N \times C \times 3}$. We learn an element-wise signal of data dependent directions $\mathcal{K} \in \sR^{N \times C \times 3}$. Similarly to \Section{nonlinearity}, these directions are obtained via applying a weight matrix $\rmW \in \sR^{C \times C}$ to each $\mV_n \in \gV$:
\begin{align}
    \mathcal{K} = \{ \rmW \mV_n \}_{n=1}^N,
\end{align}
%
%
and then computing the element of $\gV$ that best aligns with $\mathcal{K}$ and selecting it as our global feature: for each channel $c\in[C]$,
\begin{align}
    f_{\text{MAX}}(\gV)[c] &= \mV_{n^*}[c] \\
    \text{where}\quad n^*(c) &=\argmax_n \langle \rmW\mV_n[c], \mV_n[c]\rangle.
    \label{eq:pooling}
\end{align}
where $\mV_n[c]$ stands for the vector channel $\vv_c \in \mV_n$.

Similarly, we can aggregate information locally (local pooling) by grouping $k$ nearest neighbours in $\gV$ and perform the aforementioned pooling seperately for each group. 

\begin{figure}[t]
\centering
\includegraphics[width=.95\linewidth]{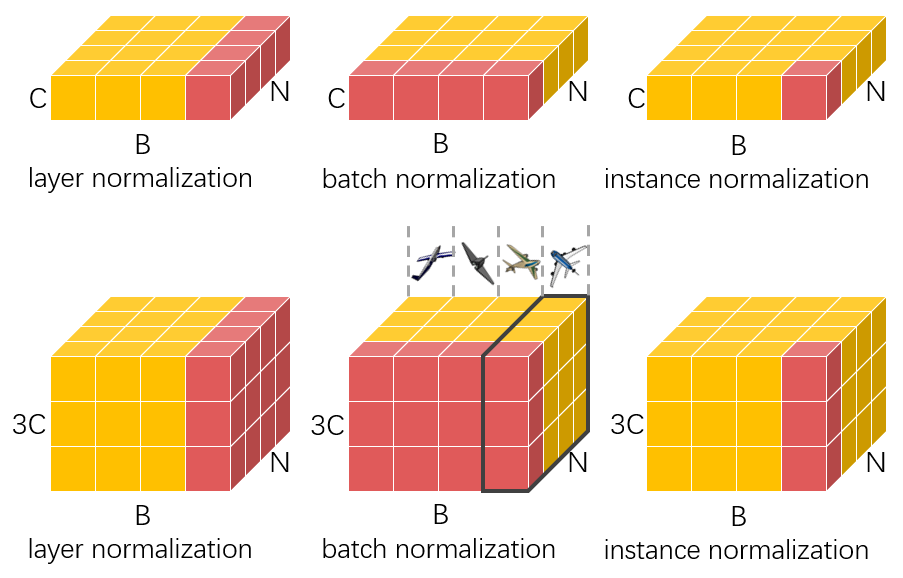}
\caption{\textbf{Normalizations --}
$\mathrm{B,C,N}$ stand for the batch, channel, and point dimensions respectively. (Top) classical scalar neurons; (bottom) vector neurons, batch normalization can only be done on vector norms as features within a batch are from different poses.}
\label{fig:normalizations}
\end{figure}
\subsection{Normalization layers -- \Figure{normalizations}}
\label{sec:normalization}
Normalization often give rise to significant performance improvements.
Layer \cite{ba2016layer} and instance normalizations \cite{ulyanov2017instance} are done pre-sample (and the latter also per channel) and thus can be trivially generalized to VN networks, where the distributions are normalized with respect to vector distributions in $\R^3$.

\paragraph{Batch normalization}
In contrast to other forms of normalizations, batch normalization aggregates statistics across all batch samples. While technically possible, in the context of rotation equivariant networks, averaging across arbitrarily rotated inputs would not necessarily be meaningful. For example, averaging two input features rotated in opposite directions would zero them out instead of producing that feature in a canonical pose.  


We instead apply batch normalization to the \textit{invariant} component of the vector-list features, by normalizing the 2-norms of the vector-list features.
%

Given a batch of $B$ vector-list features $\{\mV_b\}_{b=1}^B$ with each $\mV_b \in \sR^{C\times3}$, 
our batch normalization is defined as:
\begin{align}
    \label{eq:vn-batchnorm_norm}
    \mN_b &= \mathrm{ElementwiseNorm}(\mV_b) \in \sR^{N \times 1} \\
    %
    \{\mN'_b\}_{b=1}^B &= \mathrm{BatchNorm}\left(\{\mN_b\}_{b=1}^B\right) \\
    \mV'_b[c]
    &= \mV_b[c]\, \frac{~\mN'_b[c]~}{~\mN_b[c]~}, \quad \forall\, c\in[C], 
\end{align}
where $\mV'_b[c], \mV_b[c]$ are the vector channels, $\mN'_b[c], \mN_b[c]$ are their scalar 2-norms, and $\mathrm{ElementWiseNorm}(\mV_b)$ computes the 2-norm of every vector channel $\vv_c=\mV_b[c] \in \mV_b$. 


\subsection{Invariant layers}
\label{sec:invariant}
%

General invariant architectures are comprised of equivariant layers followed by invariant ones. We now introduce our invariant layer, that can be appended as needed to the output of the equivariant VN layers.
Rotation-invariant networks are essential for both classification and segmentation tasks, where the identity of an object or its parts should be invariant to pose.

Key to our approach is the idea that the product of an equivariant signal $\mV \in \sR^{C \times 3}$ by the transpose of an equivariant signal $\mT \in \sR^{C' \times 3}$ is rotation invariant:
\begin{equation}
    (\mV R ) (\mT R)^\top
    =
    \mV R R^\top \mT^\top
    =
    \mV \mT^\top.
    \label{eq:invar_prod}
\end{equation}
Note that a specific case of \eq{invar_prod} is the inner product of two vectors, in particular the norm of equivariant vector features is rotation invariant.

We could compute an invariant feature from a vector-list $\mV\in\R^{C\times3}$ as the Gram matrix $\mV \mV^\top$.
However, this would result in a large $O(C^2)$ storage complexity. We could also consider taking the norm of each row of $\mV$ but this would result in the loss of the relative directional information between the rows.
Instead we propose a scalable solution with more manageable $O(C)$ complexity that can preserve directional information.

Our idea is to produce a coordinate system $\mT \in \sR^{3 \times 3}$ from $\mV$ and read $\mV$ in this coordinate system thus producing rotation invariant features. In practice we consider our usual set of equivariant vector-list feature $\gV \in \sR^{N \times C \times 3}$. 
Inspired by \citet{maron2020learning}, we produce a matrix $\mT_n$ for each element by concatenating its feature $\mV_n \in \sR^{C \times 3}$ with the global mean $\overline{\mV} := \frac{1}{N} \sum_n \mV_n \in \sR^{C \times 3}$ and running through a vector neuron MLP with a target number of channel $C' = 3$:
%
\begin{equation}
    \mT_n := \text{VN-MLP}([\mV_n, \overline{\mV}])
    \label{eq:pose_matrix}
\end{equation}
Finally we define our invariant layer by:
\begin{equation}
    \text{VN-In}(\mV_n) := \mV_n \mT_n^\top.
    \label{eq:invar_layer}
\end{equation}

\section{Network Architectures}
\label{sec:architectures}

We now show how we can plug vector neurons into two widely used 3D learning architectures, PointNet \cite{qi2017pointnet} and DGCNN \cite{wang2019dynamic}.
These two backbones are representative of the richness of pointcloud networks, as PointNet is free from convolutions, and DGCNN comprises convolutions but the message passing is on dynamic graphs whose edges are not directly embedded in $\R^3$.
As we show next, VN networks fit well into these backbones, while previous convolution-based methods such as TFN \cite{thomas2018tensor} and EGCL \cite{satorras2021n} do not.
To make clear the ease of generalization, in the following we will adopt a naming convention to the functions defined in section \ref{sec:method} by using a ``VN'' prefix.

\paragraph{VN-DGCNN}
DGCNN performs a permutation equivariant edge convolution by computing adjacent edge features $\ve'_{nm}$ followed by a local max pooling:
\begin{align}
    \ve'_{nm} &= \mathrm{ReLU}(\Theta(\vx_m-\vx_n) + \Phi\vx_n) \\
    \vx_n' &= \mathrm{Pool}_{m:(n,m)\in\gE}( \ve'_{nm}),
\end{align}
where $\vx_n\in\sR^3$ are per-point features and $\Theta,\Phi$ are learnable weight matrices.
Our VN-DGCNN requires a straightforward modification:
\begin{align}
    \mE'_{nm} &= \text{VN-ReLU}(\Theta(\mV_m-\mV_n) + \Phi\mV_n) \\
    \mV_n' &= \text{VN-Pool}_{m:(n,m)\in\gE}( \mE'_{nm})
\end{align}
using our vector-list representation $\mV_n\in\sR^{C\times3}$.

\paragraph{VN-PointNet}
PointNet approximates a permutation symmetric function using
\begin{equation}
    \vx' = \mathrm{Pool}_{\vx_n\in\gX}(h(\vx_1),\cdots,h(\vx_N)),
\end{equation}
where $h$ is the same for all $\vx_n$.
Its VN version is written as
\begin{equation}
    \mV' = \text{VN-Pool}_{\mV_n\in\gV}(f(\mV_1),\cdots,f(\mV_N)),
\end{equation}
where $f$ is a shared VN-MLP.
%
One issue here exists in the first input layer where the input pointcloud coordinates $\mV_i$ are $\sR^{1\times3}$ vectors and thus applying $f$ to them would degenerate to a set of $\sR^{C\times3}$ vector-lists whose vector components are all linearly dependent (pointing to one direction).
This is analogous to applying a per-pixel 1x1 convolution to a gray-scale image (single input channel). 
Therefore, in VN-PointNet we add an edge convolution at the input layer, mapping $\sR^{1\times3}$ features into $\sR^{C\times3}$ with $C>1$ and then continue with per-point VN-MLP operations.

\section{Experiments}
\label{sec:exp}
We evaluate our method on three core tasks in pointcloud processing: classification~(\Section{exp:cls}), segmentation~(\Section{exp:seg}), and reconstruction~(\Section{exp:sdf}).
In addition to their diversity in the required output, these tasks span different use cases of our proposed equivariant framework: classification and segmentation are rotation-invariant tasks, while reconstruction is rotation-equivariant. 

\paragraph{Datasets} 
We employed the ModelNet40~\cite{shapenet2015} and the ShapeNet~\cite{shapenet2015} datasets for evaluation. The ModelNet40 dataset consists of 40 classes with 12,311 CAD models in total. We used 9,843 models for training and the others for testing in the classification task. For the ShapeNet dataset, we followed \cite{yi2016scalable} by using ShapeNet-part for part segmentation, which has 16 shape categories with more than 30,000 models. We also applied the subset of ShapeNet in \cite{choy20163d} for shape reconstruction, containing 13 major categories with~50,000 models.

\paragraph{Train/test rotation setup}
In classification and segmentation, following the conventions from~\citet{esteves2018learning}, we adopt three train$/$test settings: $z/z$, $z/\mathrm{SO}(3)$ and $\mathrm{SO}(3)/\mathrm{SO}(3)$, where $z$ stands for data augmentation with rotations only around the $z$ axis, and $\mathrm{SO}(3)$ for arbitrary rotations.
All rotations are generated on the fly at the training time, thereby comparing the equivariance-by-construction of VN architectures with a \textit{learned-by- augmentation} equivariance.
At test time, each shape is presented at a single rotation.
For reconstruction, we show results on extreme settings: no-rotation (I) -- the standard evaluation setup for prior methods, and arbitrary rotations $\mathrm{SO}(3)$.
Since outputs in this task are static and optimization for each shape takes multiple iterations at both train and test times, here the $\mathrm{SO}(3)$ random rotations are generated for each shape in a pre-processing stage and all shapes stay at fixed poses during training.

\paragraph{Network implementations}
In classification and segmentation, we implement our VN networks in the identical architectures to their classical counterparts, but with each layer in the shape of $\lfloor\frac{N}{3}\rfloor\times3$ while the corresponding layer in the scalar network has size $N$.
This in fact greatly reduces the number of learnable parameters in VN networks, resulting in roughly $\leqslant2/3^2 = 2/9$ times of parameters compared to the counterpart scalar networks -- here the factor 2 in the numerator is because in nonlinearities two components $\vq,\vk$ are both learned (Equation \ref{eq:linearEquivariance}).
In reconstruction we slightly extend the layer size for the VN encoder.
Moreover, in VN-PointNet, we discard the input spatial transformation MLP which learns $3\times3$ transformation matrices as our VN network already takes rigid transformations into consideration by construction.
In the following experiments, we use mean pooling as aggregation in all networks, which performed better in practice.
We will discuss more about the max pooling as well as ablation study on other structures in the \SupplementaryMaterial. 

\newcolumntype{R}{>{\columncolor{LightRed}}c}
\begin{table}[ht]
  \centering
  \begin{tabular}{ccRc}
    \toprule
    Methods & $z/z$ & $z/\mathrm{SO}(3)$ & $\mathrm{SO}(3)/\mathrm{SO}(3)$ \\
    \midrule
    \multicolumn{4}{c}{Point / mesh inputs} \\
    \midrule
    PointNet \cite{qi2017pointnet} & 85.9 & 19.6 & 74.7 \\
    DGCNN \cite{wang2019dynamic} & 90.3 & 33.8 & 88.6 \\
    VN-PointNet & 77.5 & 77.5 & 77.2 \\
    VN-DGCNN & 89.5 & \textbf{89.5} & \textbf{90.2} \\
    \midrule
    PCNN \cite{atzmon2018point} & 92.3 & 11.9 & 85.1  \\
    ShellNet \cite{zhang2019shellnet} & \textbf{93.1} & 19.9 & 87.8 \\
    PointNet++ \cite{qi2017pointnetpp} & 91.8  & 28.4 & 85.0 \\
    PointCNN \cite{li2018pointcnn} & 92.5 & 41.2 & 84.5 \\
    %
    Spherical-CNN \cite{esteves2018learning} & 88.9 & 76.7 & 86.9 \\
    $a^3$S-CNN \cite{liu2018deep} & 89.6 & 87.9 & 88.7 \\
    \midrule
    SFCNN \cite{rao2019spherical} & 91.4 & 84.8 & 90.1 \\
    TFN \cite{thomas2018tensor} & 88.5 & 85.3 & 87.6 \\
    RI-Conv \cite{zhang2019rotation} & 86.5 & 86.4 & 86.4 \\
    SPHNet \cite{poulenard2019effective} & 87.7 & 86.6 & 87.6 \\
    ClusterNet \cite{chen2019clusternet} & 87.1 & 87.1 & 87.1 \\
    GC-Conv \cite{zhang2020global} & 89.0 & 89.1 & 89.2 \\
    RI-Framework \cite{li2020rotation} & 89.4 & 89.4 & 89.3 \\
    \midrule
    \multicolumn{4}{c}{Point + normal inputs} \\
    \midrule
    \rowcolor[gray]{0.9} SFCNN \cite{rao2019spherical}& 92.3 &  85.3 & 91.0 \\
    \rowcolor[gray]{0.9} LGR-Net \cite{zhao2019rotation} & 90.9 & 90.9 & 91.1 \\
    \bottomrule
  \end{tabular}
  \caption{Test classification accuracy on the ModelNet40 dataset \cite{wu20153d} in three train/test scenarios. $z$ stands for aligned data augmented by random rotations around the vertical axis and $\mathrm{SO}(3)$ indicates data augmented by random rotations.
  }
  \label{tab:exp:cls}
\end{table}
\subsection{Classification -- Table~\ref{tab:exp:cls}}
\label{sec:exp:cls}
%
%
We evaluate classification results on ModelNet40 compared with vanilla PointNet, DGCNN, and other rotation invariant or equivariant methods which takes point coordinates (meshes or pointclouds) as inputs.
Compared with their non-equivariant counterparts, the VN networks attain consistently good results on all the three settings, which indicates their robustness over rotations, especially in the $z/\mathrm{SO}(3)$ case where the test set contains unseen rotations in the train set.
Even in the $\mathrm{SO}(3)/\mathrm{SO}(3)$ case with abundant train-time data augmentation, the rotation sensitive networks cannot perform as well as the equivariance by construction in VN networks.
One the other hand, our VN network with DGCNN backbone (VN-DGCNN) outperforms all other equivariant or invariant methods with only point coordinate inputs in the $z/\mathrm{SO}(3)$ and $\mathrm{SO}(3)/\mathrm{SO}(3)$ cases.
Note that methods that use surface normals \cite{rao2019spherical, zhao2019rotation} still achieve better slightly better results.
%


\newcolumntype{R}{>{\columncolor{LightRed}}c}
\begin{table}[th]
  \centering
  \begin{tabular}{cRc}
    \toprule
    Methods & $z/\mathrm{SO}(3)$ & $\mathrm{SO}(3)/\mathrm{SO}(3)$ \\
    \midrule
    \multicolumn{3}{c}{Point / mesh inputs} \\
    \midrule
    PointNet \cite{qi2017pointnet} & 38.0 & 62.3 \\
    DGCNN \cite{wang2019dynamic} & 49.3 & 78.6 \\
    VN-PointNet & 72.4 & 72.8 \\
    VN-DGCNN & \textbf{81.4} & \textbf{81.4}\\
    \midrule
    PointCNN \cite{li2018pointcnn} & 34.7 & 71.4 \\
    PointNet++ \cite{qi2017pointnetpp} & 48.3 & 76.7 \\
    ShellNet \cite{zhang2019shellnet} & 47.2 & 77.1 \\
    \midrule
    RI-Conv \cite{zhang2019rotation} & 75.3 & 75.3 \\
    TFN \cite{thomas2018tensor} & 76.8 & 76.2 \\
    GC-Conv \cite{zhang2020global} & 77.2 & 77.3 \\
    RI-Framework \cite{li2020rotation} & 79.2 & 79.4 \\
    \midrule
    \multicolumn{3}{c}{Point + normal inputs} \\
    \midrule
    \rowcolor[gray]{0.9} LGR-Net \cite{zhao2019rotation} & 80.0 & 80.1 \\
    \bottomrule
  \end{tabular}
  \caption{ShapeNet part segmentation. The results are reported in overall average category mean IoU over 16 categories in two train/test scenarios. With $z$, we refer to data augmented only by random rotations around the vertical axis, and $\mathrm{SO}(3)$ indicates random rotations.}
  \label{tab:exp:seg}
\end{table}
\subsection{Part segmentation -- Table~\ref{tab:exp:seg}}
\label{sec:exp:seg}
Table \ref{tab:exp:seg} shows our results in ShapeNet~part segmentation.
Again our method shows consistent results across different rotations and achieves best performance with VN-DGCNN compared with other works, including~\cite{zhao2019rotation} that uses surface normals in addition to the point coordinates.


\subsection{Neural implicit reconstruction -- Table~\ref{tab:exp:sdf}}
\label{sec:exp:sdf}
We follow the pointcloud \textit{completion} experiment from OccNet~\cite{mescheder2019occupancy}, where we reconstruct neural implicit functions from sparse and noisy input pointclouds: we subsample 300 points from the surface of each~(water-tight) ShapeNet model, and perturb them with normal noise with zero mean and $0.005$ standard deviation.
The outputs are occupancy probability functions $\mathcal{O}: \sR^3\to[0,1]$ which can be parameterized by a shared neural implicit function $h_\theta(\cdot\,|\,\vz): \sR^3\to[0,1]$ conditioned by a latent code~$\vz$ derived from the input point set.
For fair comparisons, we retrain the original OccNet~\cite{mescheder2019occupancy} together with our methods for 300k iterations, and select the models with the best performance on the validation set.

\paragraph{Encoder network}
We build an encoder-decoder framework with the architecture similar to \cite{mescheder2019occupancy} but in the language of VN.
The encoder is rotation equivariant, encoding a pointcloud $\{\vx_1,\vx_2,\cdots,\vx_1\}$ into a global vector-list feature $\mZ\in\sR^{C\times3}$.
While in \citet{mescheder2019occupancy} the encoder is a PointNet, here we use a VN-PointNet:
\begin{equation}
    \mZ = \text{VN-PointNet}(\{\vx_1,\vx_2,\cdots,\vx_1\}).
\end{equation}

\paragraph{Decoder network}
The decoder is rotation invariant between vector-list latent code $\mZ\in\sR^{C\times3}$ and query point coordinate $\vx\in\sR^3$ -- if the shape and the query point are simultaneously rotated, the occupancy value stays unchanged.
We define the decoder as a function on the three invariant features $\|\vx\|^2, \langle \vx,\mZ\rangle, \text{VN-In}(\mZ)$:
\begin{equation}
    \mathcal{O}(\vx\,|\,\mZ) = \text{ResNet}([\langle \vx, \mZ\rangle), \|\vx\|^2, \text{VN-In}(\mZ)]),
\end{equation}
where $\text{VN-In}(\cdot)$ is the VN invariant layer defined in~\Section{invariant}.
As an ablation study, we also replace VN-PointNet with a standard PointNet encoder (with the same invariant decoder), where the encoder generates latent codes $\vz\in\sR^C$ and we reshape them into $\mZ\in\sR^{(C/3)\times3}$.

%
On the contrary, the decoder in \cite{mescheder2019occupancy} is a simple non-linear function $\mathcal{O}(\vx\,|\,\vz)= h(\varphi(\vx),\psi(\vz))$, which given latent code $\vz\in\sR^C$ and query point $\vx\in\sR^3$ outputs an occupancy probability $\mathcal{O} \in [0,1]$
\footnote{\cite{mescheder2019occupancy} provides multiple versions of decoders. We select this simplest one in our experiments for easier comparisons.}.
%
%
%

\newcolumntype{R}{>{\columncolor{LightRed}}c}
\begin{table*}[th]
  \centering
  \begin{tabular}{c|ccc|cRcc}
    \toprule
    Methods & Encoder & Latent code & Decoder & $\mathrm{I}/\mathrm{I}$ & $\mathrm{I}/\mathrm{SO}(3)$ & $\mathrm{SO}(3)/\mathrm{SO}(3)$ \\
    \midrule
    OccNet \cite{mescheder2019occupancy} & PointNet & $\vz\in\sR^{C}$ & $h(\varphi(\vx),\psi(\vz))$ &
    71.4 & 30.9 & 58.2 \\
    \midrule
    - & PointNet & $\mZ\in\sR^{\lfloor C/3\rfloor\times3}$ & $h(\langle \vx,\mZ\rangle, \|\vx\|^2, \text{VN-In}(\mZ))$ &
    \textbf{72.0} & 31.0 & 59.4 \\
    VN-OccNet & VN-PointNet & $\mZ\in\sR^{\lfloor C/3\rfloor\times3}$ & $h(\langle \vx,\mZ \rangle, \|\vx\|^2, \text{VN-In}(\mZ))$ &
    69.3 & \textbf{69.3} & \textbf{68.8} \\
    \bottomrule
  \end{tabular}
  \caption{Volumetric mIoU on ShapeNet reconstruction with neural implicits. We show results on extreme settings: no-rotation (I) -- the standard evaluation setup for prior methods, and arbitrary rotations $\mathrm{SO}(3)$. Here the $\mathrm{SO}(3)$ random rotations are generated for each shape in a pre-processing stage and all shapes stay at fixed poses during training.}
  \label{tab:exp:sdf}
\end{table*}
\begin{figure*}[t]
    \centering
    \includegraphics[clip, trim=3cm 0 3cm 0, width=\linewidth]{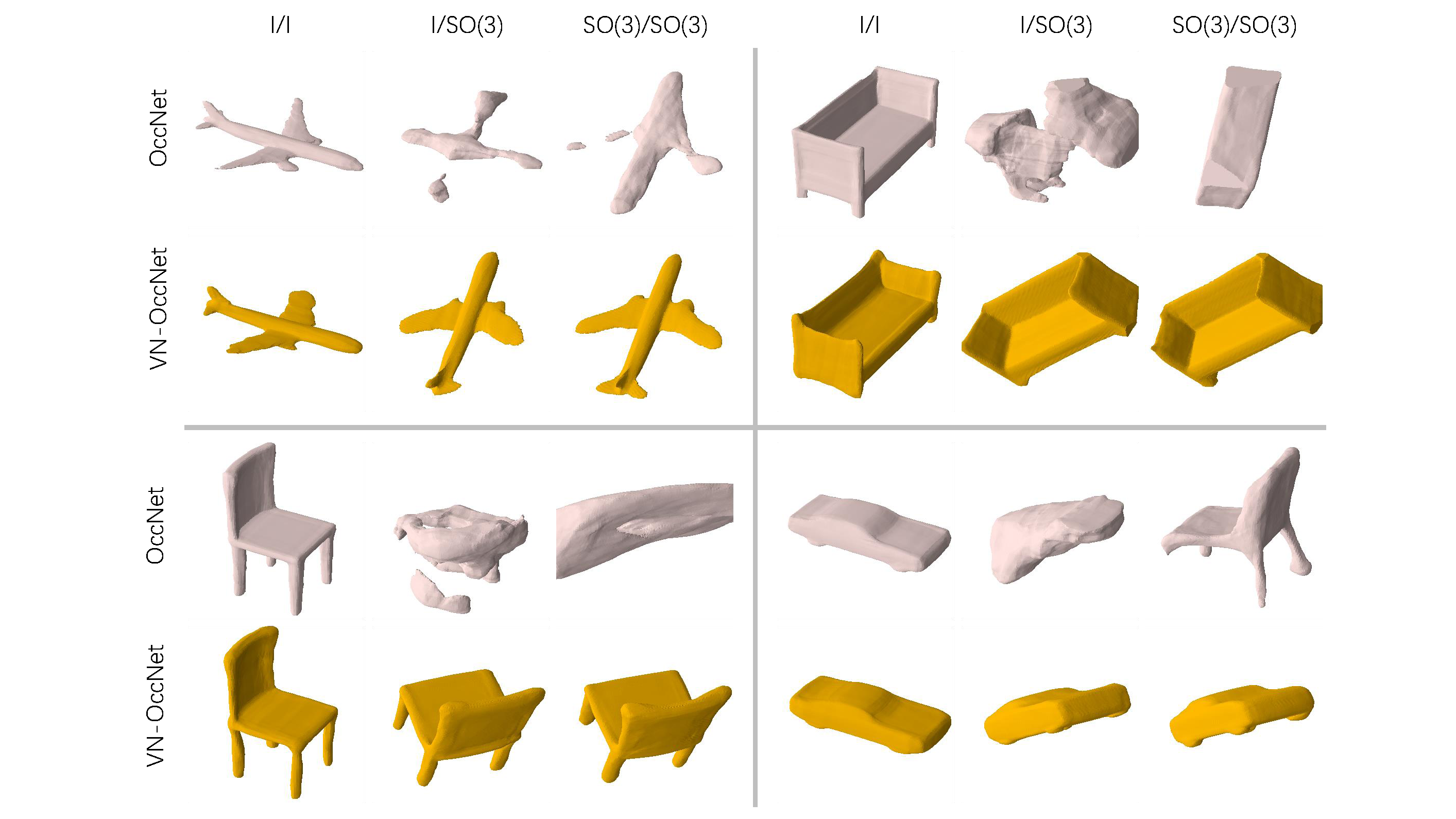}
    \caption{Reconstruction results on ShapeNet with OccNet ({\color{mesh_misty_rose}light pink}) and VN-OccNet ({\color{mesh_yellow}yellow}). Meshes are extracted from the neural implicits using the Multi-resolution IsoSurface Extraction~(MISE) method.}
    \label{fig:reconstruction_vis}
\end{figure*}
\paragraph{Quantitative results -- Table \ref{tab:exp:sdf}}
We evaluate the volumetric mean IoU of the reconstructions in the three train/test settings.
Compared with the original OccNet, our equivariant-encoder/invariant-decoder exhibits excellent coherence in reconstructing shapes in any poses, with a minor loss in accuracy in the $\mathrm{I}/\mathrm{I}$ case.
Even simply adopting the invariant decoder without an equivariant encoder slightly improves the performances in all the three settings. 

\paragraph{Qualitative results -- \Figure{reconstruction_vis}}
We show some reconstructions from the test set using the original OccNet ({\color{mesh_misty_rose}light pink}) and our VN-OccNet ({\color{mesh_yellow}yellow}).
Meshes are extracted from the neural implicits using the Multi-resolution IsoSurface Extraction~(MISE) method from~\citet{mescheder2019occupancy}.
Although OccNet better recognizes the fine details in the $\mathrm{I}/\mathrm{I}$ case when the dataset is pre-aligned, it is extremely sensitive to rotations.
In the $\mathrm{I}/\mathrm{SO}(3)$ case when unseen rotations are applied at test time, OccNet totally fails by hardly learning anything meaningful; these findings are also consistent with those in~\citet{deng2020nasa}.
Even in the $\mathrm{SO}(3)/\mathrm{SO}(3)$ case when data augmentation is adopted at train time, it still shows its limitation by generating blurry shapes (top left), averaged shapes (top right, the box-like output consists of sofa features averaged from different poses), or shapes with incorrect priors~(bottom right, a shape in the car class is falsely identified as a chair).

\section{Conclusions}

We have introduced Vector Neurons -- a novel framework that facilitates rotation equivariant neural networks by lifting standard neural network representations to 3 space. To that end, we have introduced the vector-neuron counterpart of standard network modules: linear layers, non-linearities, pooling and normalization.
Using our framework, we have built a rotation-equivariant version of two leading pointcloud network backbones: PointNet and DGCNN, and evaluated them on 3 tasks: classification, segmentation and reconstruction.
Our results demonstrate a consistent advantage to our modified architecture when the input shapes pose is arbitrary, compared to an augmentation based approach. 

\paragraph{Limitation and future work}
While our method shines under arbitrary rotation settings, on aligned input shapes and specifically in the task of reconstruction, our VN-OccNet was not able to match the reconstruction quality of vanilla OccNet by a small margin.
In future work we plan to investigate this matter.

In this work, we have focused on 3D pointcloud networks, yielding permutation and rotation equivariant architectures. However, it should be clear that our framework has obvious generalizations to higher-dimensional pointclouds in a completely analogous way. We also believed it can find applications in other modalities like meshes, voxel grids, and even in the image domain. Generalization of vector neurons to other transformation groups of interest, such as the full affine group, can also be investigated (the addition of uniform scalings in our framework is quite straightforward).

In summary, by making rotation equivariant modules simple and accessible we hope to alleviate the need to curate and pre-align shapes for supervision and inspire future research on this fascinating topic.

\paragraph{Acknowledgements}
We gratefully acknowledge the support of a Vannevar Bush Faculty Fellowship, as well as gifts from the Adobe, Amazon AWS, and Autodesk corporations.



{\small
\bibliographystyle{plainnat}
\bibliography{macros,egbib}
}

\clearpage
\twocolumn[
\centering
\Large
\textbf{Vector Neurons: A General Framework for SO(3)-Equivariant Networks} \\
\vspace{0.5em}
(Supplementary Material) \\
\vspace{1.0em}
]


\section{Discussions}

In this section, we discuss some extensions, alternatives, and explanations to the VN layers in Section \ref{sec:method}.

\subsection{Non-linearity}
\label{sec:suppl:nonlinearity}
\begin{figure}[h]
\centering
\includegraphics[width=.85\linewidth]{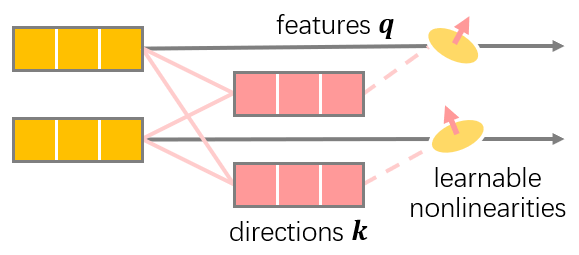}
\caption{
A detached non-linear layer without built-in linear layer.
}
\label{fig:nonlinearity_overall_separate}
\end{figure}

\paragraph{Linear and non-linear layers -- \Fig{nonlinearity_overall_separate}}
The VN-ReLU defined in Section \ref{sec:nonlinearity} already consists of a built-in linear layer $\vq = \rmW\mV$ \eq{linearEquivariance} and the the non-linearity is applied to this learned feature $\vq$.
An alternative to this is to construct linear and non-linear layers separately, where the non-linearity is directly applied to each input vector channel $\vv\in\mV$ by
\begin{equation}
    \vv'
    = \begin{cases}
        \vv &
        \text{if\ } \langle \vv,\vk \rangle \geqslant 0 \\
        \vv - \left\langle \vv,\frac{\vk}{\|\vk\|} \right\rangle \frac{\vk}{\|\vk\|} &
        \text{otherwise,}
    \end{cases}
    \label{eq:VN-ReLU_detached}
\end{equation}
Detaching the linear layer from non-linearity allows more flexibility in constructing neural networks and, in practice, gives better results in some cases.
However, this also doubles the network depth and can lead to longer training time compared to the entangled linear-ReLU layer in \eq{VN-ReLU}.
Experimental comparisons will be shown in Section \ref{sec:suppl:ablation}.

\paragraph{Other non-linearities}
Though we only showed how to define VN-ReLU in Section \ref{sec:nonlinearity}, a rich library of equivariant non-linearities can be defined in this manner using the input-dependent direction vector $\vk$.
An immediate extension is VN-LeakyReLU, where instead of clipping $\vq_\parallel$ to zero we contract it by a factor $\alpha\in(0,1)$.
In the manner of the detached VN-ReLU in \eq{VN-ReLU_detached}, the VN-LeakyReLU can be easily expressed as:
\begin{equation}
    f_{\text{LeakyReLU}}(\mV;\alpha) = \alpha \mV + (1-\alpha) f_{\text{ReLU}} (\mV).
\end{equation}
An entangled layer of VN-Linear and VN-LeakyReLU can also be defined analogous to \eq{VN-ReLU}.
More generally, given an arbitrary non-linear scalar function $h:\sR\to\sR$, we can incorporate it into our VN non-linearity framework by applying it to $\vq_\parallel$ along the $\vk$ direction, namely,
\begin{equation}
    \vv' = \frac{h(\|\vq_\parallel\|)}{\|\vq_\parallel\|}~ \vq_\parallel + \vq_\perp.
\end{equation}

\subsection{Local Pooling}
The VN-MAX pooling in Section \ref{sec:pooling} is defined across an entire pointcloud $\gV \in \sR^{N \times C \times 3}$, but we can also aggregate information locally via local pooling.

\paragraph{In the primal space}
For any point $\vx\in\gX$ with feature $\mV\in\gV$ we consider its $K$ nearest neighbours $\{\vx_k\}_{k=1}^K$ in the primal space and we denote by $\mV_k\in\gV$ the corresponding feature of $\vx_k$.
Similar to global pooling \eq{pooling}, local pooling (in the primal space) is given by:
\begin{align}
    f_{\text{MAX}}\left(\{\mV_k\}_{k=1}^K\right)[c] &= \mV_{k^*}[c] \\
    \text{where}\quad k^*(c) &= \argmax_k \: \langle \rmW_k\mV_k[c], \mV_k[c]\rangle.
    \label{eq:pooling_local}
\end{align}

\paragraph{Feature space locality}
As in DGCNN \cite{wang2019dynamic}, we can also query the $K$ nearest neighbours $\{\mV_k\}_{k=1}^K$ of feature $\mV_n\in\gV$ in the feature space $\sR^{C\times3}$ directly, followed by local pooling (in the feature space) with exactly the same formulation as \eq{pooling_local}.

\subsection{Batch Normalization}
In VN-BatchNorm \eq{vn-batchnorm_norm}, for each input vector-list feature $\mV_b$, all entries in its per-channel 2-norm $\mN_b$ are non-negative, but after normalizing the distributions, the output ``2-norm'' $\mN'_b$ can have negative entries.
Geometrically, a negative entry $\vn'_c \in \mN'_b$ means the orientation of its corresponding vector channel is flipped, that is, $\vv'_c\in\mV'_b$ is in the opposite direction of $\vv_c\in\mV_b$.

To avoid the negative 2-norms, an alternative is to take logarithms on all entries of $\mN_b$ and then apply the standard BatchNorm to $\log(\mN_b)$.
So the VN batch normalization becomes:
\begin{align}
    \label{eq:vn-batchnorm_lognorm}
    \mN_b &= \mathrm{ElementWiseNorm}(\mV_b) \in \sR^{N \times 1} \\
    \{\mN'_b\}_{b=1}^B &= \mathrm{BatchNorm}\left(\{\log(\mN_b)\}_{b=1}^B\right) \\
    \mV'_b[c]
    &= \mV_b[c]\, \frac{\exp(\mN'_b[c])}{\mN_b[c]},
\end{align}
where $\log$ and $\exp$ act element-wise.
However, taking $\log$ and $\exp$ brings a lot of instability and in practice can cause gradient explosion. Also, logarithms cannot be computed for vectors with zero 2-norms.


\section{Additional Experiments}

\newcolumntype{R}{>{\columncolor{LightRed}}c}
\begin{table}[h]
  \centering
  \begin{tabular}{c|ccR}
    \toprule
    Method & $\mathrm{I}/\mathrm{I}$ & $\mathrm{I}/z$ & $\mathrm{I}/\mathrm{SO}(3)$ \\
    \midrule
    PointNet & 90.7 & 23.1 & 7.9 \\
    DGCNN & \textbf{92.9} & 37.2 & 16.6  \\
    VN-PointNet & 77.2 & 77.2 & 77.2 \\
    VN-DGCNN & 90.0 & \textbf{90.0} & \textbf{90.0} \\
    \bottomrule
  \end{tabular}
  \caption{Test classification accuracy (\%) on the ModelNet40 dataset \cite{wu20153d} with training on aligned data. $\mathrm{I}$ stands for no-rotations.}
  \label{tab:exp-suppl:rotations_cls}
\end{table}
\newcolumntype{R}{>{\columncolor{LightRed}}c}
\begin{table}[h]
  \centering
  \begin{tabular}{c|ccR}
    \toprule
    Method & $\mathrm{I}/\mathrm{I}$ & $\mathrm{I}/z$ & $\mathrm{I}/\mathrm{SO}(3)$ \\
    \midrule
    PointNet & 78.7 & 36.7 & 30.3 \\
    DGCNN & \textbf{85.2} & 43.8 & 36.1  \\
    VN-PointNet & 73.0 & 73.0 & 73.0 \\
    VN-DGCNN & 81.5 & \textbf{81.5} & \textbf{81.5} \\
    \bottomrule
  \end{tabular}
  \caption{ShapeNet part segmentation results (mIoU). Training is done on aligned data without rotation augmentation.}
  \label{tab:exp-suppl:rotations_seg}
\end{table}

\newcolumntype{R}{>{\columncolor{LightRed}}c}
\begin{table}[h]
  \centering
  \begin{tabular}{c|cRc}
    \toprule
    Non-lin & $z/z$ & $z/\mathrm{SO}(3)$ & $\mathrm{SO}(3)/\mathrm{SO}(3)$ \\
    \midrule
    \multicolumn{4}{c}{VN-PointNet} \\
    \midrule
    Built-in & 77.5 & 77.5 & \textbf{77.2} \\
    Detached & \textbf{78.2} & \textbf{78.1} & 76.8  \\
    \midrule
    \multicolumn{4}{c}{VN-DGCNN} \\
    \midrule
    Built-in  & 89.5 & 89.5 & \textbf{90.2} \\
    Detached & \textbf{90.8} & \textbf{90.7} & \textbf{90.2} \\
    \bottomrule
  \end{tabular}
  \caption{\textbf{Non-linearity --}
We compare the performances of entangled linear-ReLU (or linear-LeakyReLU) layers in \eq{VN-ReLU} with 2-tuples of a linear layer plus a separate non-linearity in \eq{VN-ReLU_detached}.
``Built-in'' stands for non-linearities with built-in linear transformations, while ``detached'' stands for tuples of detached linear and non-linear layers in \eq{VN-ReLU_detached}.
In most cases, with either the VN-PointNet or the VN-DGCNN backbone, disentangling linear and non-linear layers leads to slightly better results.
But this is also at the cost of a doubled network depth and a longer training time (roughly $\geqslant 1.5$ times to the entangled versions). }
  \label{tab:exp-suppl:nonlinearity}
\end{table}
\newcolumntype{R}{>{\columncolor{LightRed}}c}
\begin{table}[t]
  \centering
  \begin{tabular}{c|cRc}
    \toprule
    Pooling & $z/z$ & $z/\mathrm{SO}(3)$ & $\mathrm{SO}(3)/\mathrm{SO}(3)$ \\
    \midrule
    \multicolumn{4}{c}{VN-PointNet} \\
    \midrule
    VN-MAX & 76.7 & 76.7 & \textbf{77.7} \\
    MEAN & \textbf{77.5} & \textbf{77.5} & 77.2 \\
    \midrule
    \multicolumn{4}{c}{VN-DGCNN} \\
    \midrule
    VN-MAX & 88.9 & 89.0 & 88.6 \\
    MEAN & \textbf{89.5} & \textbf{89.5} & \textbf{90.2} \\
    \bottomrule
  \end{tabular}
  \caption{\textbf{Mean and max pooling --}
Comparisons between the VN-MAX aggregation defined in Section \ref{sec:pooling} and the standard mean aggregation (MEAN) which naturally preserves equivariance.
The two aggregations give comparable results, while MEAN pooling performs slightly better than VN-MAX in more cases.
Note that VN-MAX also introduces additional learnable weights compared to the mean aggregation.}
  \label{tab:exp-suppl:pooling}
\end{table}
\newcolumntype{R}{>{\columncolor{LightRed}}c}
\begin{table}[t]
  \centering
  \begin{tabular}{c|cRc}
    \toprule
    VN-In & $z/z$ & $z/\mathrm{SO}(3)$ & $\mathrm{SO}(3)/\mathrm{SO}(3)$ \\
    \midrule
    \multicolumn{4}{c}{VN-PointNet} \\
    \midrule
    VN-lin & 75.7 & 75.8 & 75.3 \\
    VN-lin + $\overline{\mV}$ & 77.1 & 77.2 & 76.7 \\
    VN-MLP & \textbf{78.0} & \textbf{77.8} & \textbf{77.3} \\
    VN-MLP + $\overline{\mV}$ & 77.5 & 77.5 & 77.2  \\
    \midrule
    \multicolumn{4}{c}{VN-DGCNN} \\
    \midrule
    VN-lin & 88.8 & 88.8 & 89.8  \\
    VN-lin + $\overline{\mV}$ & 89.7 & 89.7 & 89.7 \\
    VN-MLP & \textbf{89.9} & \textbf{89.9} & 90.1 \\
    VN-MLP + $\overline{\mV}$ & 89.5 & 89.5 & \textbf{90.2} \\
    \bottomrule
  \end{tabular}
  \caption{\textbf{Invariance --}
Table \ref{tab:exp-suppl:invariance} shows our ablation study on the invariant layer (VN-In) in Section \ref{sec:invariant}.
Specifically, in computing the equivariant coordinate systems $\mT_n$ following \eq{invar_prod}, we compare the combinations of the following options: whether or not concatenating the global mean $\overline{\mV}$ to the local feature $\mV$, and whether the VN-MLP is a 3-layer VN-MLP (VN-MLP) or a single VN linear layer (VN-lin).
Improvements in performance with both the global mean $\overline{\mV}$ and the 3-layer VN-MLP are minor.}
  \label{tab:exp-suppl:invariance}
\end{table}

\subsection{Training on Aligned Data}
In Section \ref{sec:exp}, we adopt the three train/test settings $z/z$, $z/\mathrm{SO}(3)$, $\mathrm{SO}(3)/\mathrm{SO}(3)$ from prior works to standardize the comparisons between different methods.
However, it is also interesting to see how each method performs when trained without any augmentation (no-rotation setting $\mathrm{I}$) but tested on rotated shapes.
Our additional results in classification and part segmentation on $\mathrm{I}/\mathrm{I}$, $\mathrm{I}/z$, and $\mathrm{I}/\mathrm{SO}(3)$ are shown in Table \ref{tab:exp-suppl:rotations_cls} and Table \ref{tab:exp-suppl:rotations_seg} respectively.
Compared to the $z$-trained settings in Table \ref{tab:exp:cls} and Table \ref{tab:exp:seg}, results here further highlight the robustness of our VN networks on test-time rotations in contrast to their rotation-sensitive counterparts.

\subsection{Ablation Studies}
\label{sec:suppl:ablation}
Table \ref{tab:exp-suppl:nonlinearity}, \ref{tab:exp-suppl:pooling}, and \ref{tab:exp-suppl:invariance} show our ablation studies on non-linearity, pooling, and the invariant layer in VN networks on ModelNet40 classification.
%




\end{document}